\documentclass[conference, compsocconf]{IEEEtran}
\IEEEoverridecommandlockouts
\usepackage{cite}
\usepackage{amsmath,amssymb,amsfonts}
\usepackage{algorithmic}
\usepackage{hyperref}
\usepackage{bbm}
\usepackage{graphicx}
\graphicspath{{../pdf/}{../jpeg/}{figures}}
\usepackage{textcomp}
\usepackage{xcolor}
\usepackage{subfigure}
\def\BibTeX{{\rm B\kern-.05em{\sc i\kern-.025em b}\kern-.08em
    T\kern-.1667em\lower.7ex\hbox{E}\kern-.125emX}}
    
\usepackage[letterpaper,top=72pt,bottom=54pt,left=54pt,right=54pt]{geometry}
\begin{document}
\newcommand{\ljx}[1]{\textcolor{magenta}{ljx: #1}}
\newcommand{\zhcao}[1]{\textcolor{blue}{zhcao: #1}}

\newgeometry{top=72pt,bottom=54pt,left=54pt,right=54pt}
% \title{Making Data-driven Autonomous Vehicles Governed by Digital Traffic Laws}
\title{
%Making Data-driven Autonomous Vehicles Governed by Digital Traffic Laws
Road Traffic Law Adaptive Decision-making \\ for Self-Driving Vehicles
}

\author{Jiaxin~Liu\textsuperscript{1},~Wenhui~Zhou\textsuperscript{2},~Hong~Wang\textsuperscript{1*},~Zhong~Cao\textsuperscript{1*}
\\
~Wenhao~Yu\textsuperscript{1},~Chengxiang~Zhao\textsuperscript{3},~Ding~Zhao\textsuperscript{4},~Diange~Yang\textsuperscript{1},~Jun~Li\textsuperscript{1}

\thanks{\textsuperscript{1} School of Vehicle and Mobility, Tsinghua University, Beijing, China, 100084. \textit{Email: liu-jx21@mails.tsinghua.edu.cn, \{hong\_wang,  caozhong, wenhaoyu, ydg, lijun1958\}@tsinghua.edu.cn, respectively}}
\thanks{\textsuperscript{2} Road Traffic Safety Research Center, Beijing, China, 100062, \textit{Email: 1983zhouwenhui@163.com}.}
\thanks{\textsuperscript{3} School of Mechanical Engineering, Beijing Institute of Technology, Beijing, China, 100081, \textit{Email: 3220210327@bit.edu.cn}.}
\thanks{\textsuperscript{4} Department of Mechanical Engineering, Carnegie Mellon University, USA. \textit{Email: dingzhao@cmu.edu}.}
\thanks{The corresponding authors are Zhong Cao and Hong Wang.}
}

\maketitle

% \begin{abstract
% Autonomous Vehicles (SV) should always keep compliant with traffic laws, even when it is changed by the governments. 
% However, for Reinforcement Learning (RL) based SV planning system, adapting new laws usually requires re-training RL model, which may lead to extra engineering burden and a long time unavailability of SVs. 
% This work focuses on law constraints on vehicle actions.
% A hybrid RL decision-making method is proposed for law-adaptive autonomous driving.
% In this method, the variable traffic law constraints are explicitly considered out of the RL agent to release the burden of law adaption for RL model.
% A typical well-trained RL planner is used as the backbone which makes SV decision normally. 
% The RL action will be carefully checked with the traffic laws by a law-adaptive monitor.
% Finally, a rule-based law-adaptive backup policy will intervene to make the SV action law-compliant if required.
% The method is validated in a urban driving scenario built in CARLA simulator. 
% The results show that by adapting this method, the law-breaking actions are successfully prevented and the SV can pass through the scenario with constant compliance with the laws.
% Besides, the SV can easily adapt to law change without extra engineering burden.
% \end{abstract}
\begin{abstract}

Self-driving vehicles have their own intelligence to drive on open roads. However, vehicle managers, e.g., government or industrial companies, still need a way to tell these self-driving vehicles what behaviors are encouraged or forbidden. Unlike human drivers, current self-driving vehicles cannot understand the traffic laws, and thus rely on the programmers manually writing the corresponding principles into the driving systems. It would be less efficient and hard to adapt some temporary traffic laws, especially when the vehicles use data-driven decision-making algorithms. Besides, current self-driving vehicle systems rarely take traffic law modification into consideration. This work aims to design a road traffic law adaptive decision-making method. The decision-making algorithm is designed based on reinforcement learning, in which the traffic rules are usually implicitly coded in deep neural networks. The main idea is to supply the adaptability to traffic laws of self-driving vehicles by a law-adaptive backup policy. In this work, the natural language-based traffic laws are first translated into a logical expression by the Linear Temporal Logic method. Then, the system will try to monitor in advance whether the self-driving vehicle may break the traffic laws by designing a long-term RL action space. Finally, a sample-based planning method will re-plan the trajectory when the vehicle may break the traffic rules. The method is validated in a Beijing Winter Olympic Lane scenario and an overtaking case, built in CARLA simulator. The results show that by adopting this method, self-driving vehicles can comply with new issued or updated traffic laws effectively. This method helps self-driving vehicles governed by digital traffic laws, which is necessary for the wide adoption of autonomous driving.
\end{abstract}


\section{Introduction}
% \IEEEPARstart{S}{elf-driving}
Self-driving vehicles (SVs) have their own intelligence to drive on open roads, and are widely researched for their improved traffic efficiency, safety and liberation of drivers from driving tasks\cite{wang2021risk,cao2020highway}.
The SVs should be always governed by vehicle managers for safety, social efficiency and emergency management. 
For example, exclusive lanes for public events and emergency road closures require temporary traffic control.
A common way to govern SVs is to issue new traffic laws or modify existing laws, which requires the adaptability of SVs to law variation.
Besides, the difference in traffic laws in different regions also presents challenges to SVs.

\begin{figure}[!t]
    \centering
    \includegraphics[width=0.99\linewidth]{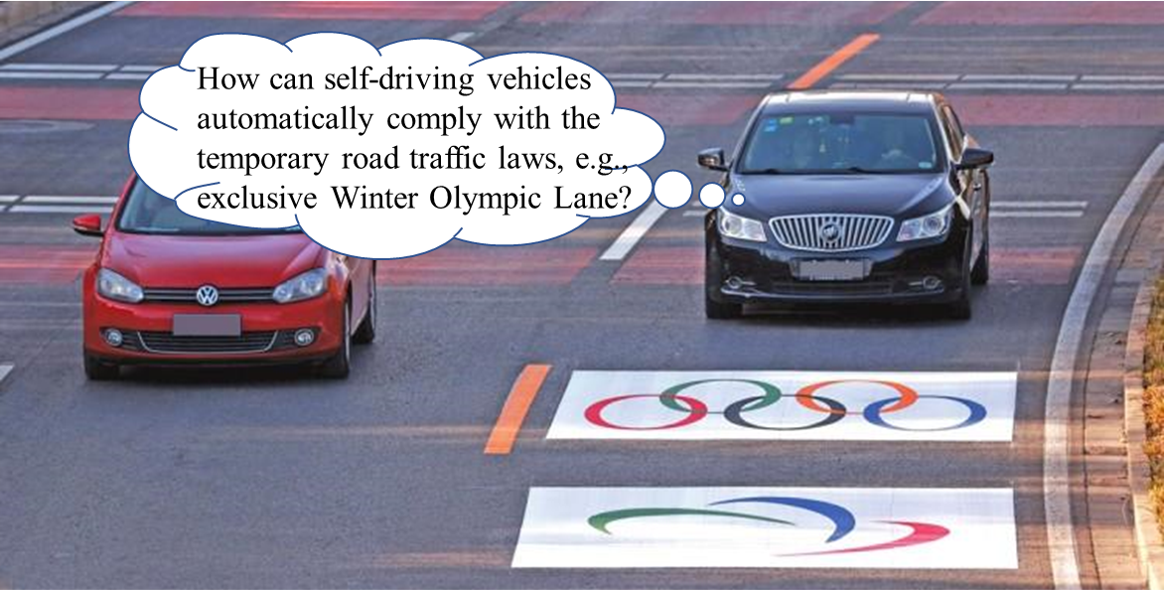}
    \caption{The self-driving vehicles should be able to adapt to new-issued or updated laws.}
\end{figure}

However, unlike human drivers, for self-driving vehicles, especially when containing data-driven deep learning algorithms, the black-box characteristic of neural networks presents challenges for SVs governing.
Since the knowledge of traffic laws is usually coded in deep neural network models implicitly, refining the model parameters for new or updated traffic rules can be intractable, while re-training the model for each version can cause unacceptable costs and the SVs may shuttle down for a long time waiting for the training.
Besides, re-training the model can be worthless for temporary traffic control.

Therefore, focusing on SVs with the deep reinforcement learning (RL) method for decision-making, our motivation is to design a law-adaptive reinforcement learning-based framework that can adapt to traffic law changes easily. 
Through this method, SVs can be effectively governed by the authority of traffic law modification.

Existing ways to consider traffic laws in reinforcement learning can be divided into training with traffic laws, building laws in state representation
%, action space pruning 
and hierarchical RL structures.

A common way to consider traffic laws is to train a policy with traffic laws by designing them as constraints or into reward function.
By formulating the traffic laws as constraints, the law-aware policy can be trained through solving a constrained Markov Decision Making problem by e.g.,  Lagrangian Multiplier-based methods\cite{tessler2018reward,liu2020ipo} , constrained policy optimization methods\cite{achiam2017constrained,schulman2015trust,chow2019lyapunov}, etc.
Besides, punishing law-violent actions in reward function is another common way\cite{talamini2020impact}. 
To maximize the expected reward, the trained RL policy tends to avoid performing law-violent actions.
However, these methods focus only on invariant laws, and as aforementioned, the implicit knowledge of laws in neural networks in these methods makes it difficult to adapt to traffic law modification.

% \restoregeometry

An intuitive way to solve the law-adaption problem is to build law constraints in state representation\cite{kuvsic2018comparison}. The variable traffic laws are modeled as part of the input of the RL agent. 
But it requires abundant training data under different laws to ensure the agent's performance, and only laws with special forms are available.
The increase of the state space dimension also brings difficulty to policy training.
Besides, this method cannot deal with newly issued traffic laws.

% Pruning actions that break the traffic laws in action space is another method to consider traffic laws in self-driving vehicles\cite{cao2020highway}.
% Using value-based reinforcement learning, this method first estimates the value of each action in action space through a deep Q value function. 
% The unsafe or law-violent actions are then pruned.
% Action with maximum value in the rest available action space will be performed.
% However, since the action space is prune in the future, the value estimation in current timestamp is inaccurate.

Rong et al.\cite{rong2020safe} proposed a hierarchical structure to ensure the safety and law compliance of SV.
In this method, a high-level RL agent outputs driving policy, based on which a rule-based low-level planner will generate a trajectory with minimum violence to safety and traffic laws.
The RL agent is trained according to the performance of the trajectory.
When traffic law changes, the low-level planner can be easily refined to keep minimum violence to traffic laws. 
But due to the hierarchical structure, when the policy generated by the high-level RL agent breaks the law, the final trajectory of the system may still not comply. 

Compared to neural network-based planners, the explainable rule-based decision-making policy can usually adapt to variable laws well, for the law constraints are usually explicitly contained in the policy and can be easily changed without re-training, e.g., Model Predictive Control based decision-making\cite{yang2021comparative,chen2020implementation,9815528}.
However, the rule-based policies usually lack optimum compared to a well-trained reinforcement learning policy.
Thus, our main idea is to supply the law-adaptive defect of the RL agent with a rule-based backup policy. 
Thus, the optimum of SV can be maintained by the RL agent, and the adaptability to traffic laws can be provided by the opportune invention of the backup policy.
The detailed contribution of this work includes,

% digitized    
% rule-based   adaptive policy and optimized policy
% \ljx{exchange?}
% adaption optimization        separate
1) A road traffic law digitization method, which converts laws described in natural languages into mathematically formulated constraints on Markov Decision Process.

2) A law-violence forecaster, which identifies actions that may break the laws in advance, and evoke the backup policy at appropriate opportunity to keep SVs legal.

3) A law-adaptive hybrid reinforcement learning decision-making framework for SVs adaption to law changes, consisting of an RL policy and a rule-based backup policy.

The remainder of this paper is organized as follows: the traffic law-adaptive decision-making problem is mathematically defined in Sec.\ref{secpb}. 
The method of traffic law digitization is introduced in Sec.\ref{seclawdig}. 
Sec.\ref{secladm} describes the detail of the proposed method, including the forecaster, long-term action space and the two policies. 
The case study is presented in Sec.\ref{case} and the paper is concluded in Sec.\ref{seccon}.

\section{Problem Definition}
\label{secpb}
In this section, the traffic law-adaptive decision-making problem is first formulated.
Then the law-adaptive hybrid RL decision-making framework is introduced.

\subsection{Road Traffic Law-Adaptive Decision Making}
This work aims to solve the following problem: \textit{a self-driving vehicle with RL-based decision-making drives in an urban scenario with traffic laws. The traffic law can be changed occasionally, and new laws can be issued. The vehicle should always keep compliant with the traffic laws even if they are modified.} 
The proposed method aims to identify the law-violent RL actions in advance adaptively, and a backup policy will intervene to control the vehicle if the RL action is not permitted.

The decision-making process can be described as,
\begin{equation}
    a^*_t = \pi_h(s_t, \mathcal{L})
\end{equation}
where $a^*$ denotes the law-compliant action at timestamp $t$. 
$\pi_h$ indicates the proposed decision-making method, which generates action $a^*$ with current state $s_t$ and a set of laws $\mathcal{L}=\left\{L_1, L_2, \dots\right\}$ to be obeyed. 
The laws are described in natural language.

In this work, instead of law sets, only single law that is to be changed or new issued is focused on.
But the proposed framework can be extended to multiple laws by increasing the dimension of state space to contain more required information.
% The law-adaptive decision making process with one law can be formulated as,
% \begin{equation}
%     a^*_t = \pi_h(s_t, L)
% \end{equation}

\subsection{Law-adaptive Hybrid RL Decision-Making Framework}

\begin{figure*}[thbp]
    \centering
    \includegraphics[width=0.99\linewidth]{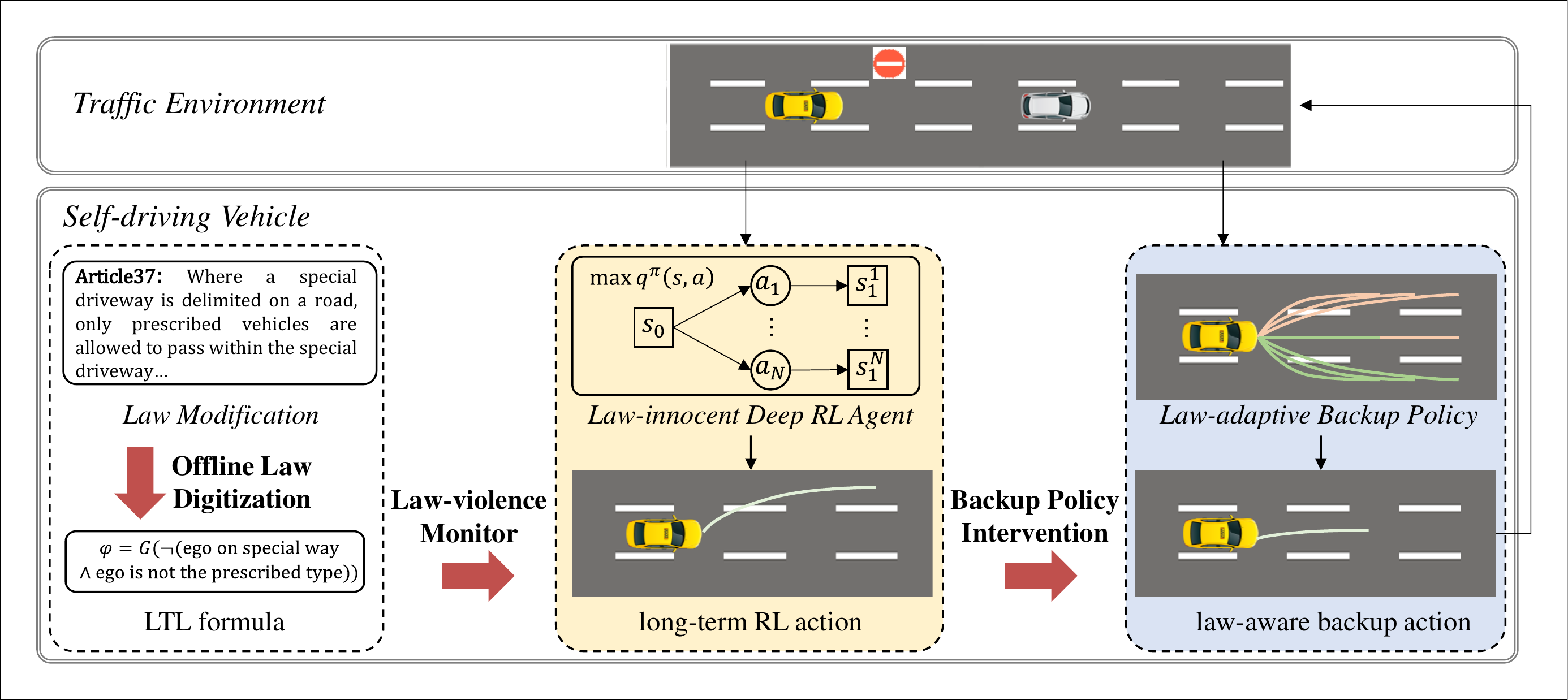}
    \caption{Traffic law adaptive decision-making framework.}
    \label{framework}
\end{figure*}

In this work, a law-adaptive hybrid reinforcement learning framework is proposed, as shown in Fig.\ref{framework}.
This method consists of a law-violence forecaster, a long-term law-innocent RL agent and a law-adaptive sample-based backup policy.
In this method, when the traffic laws change, the traffic law digitization process will be first conducted offline to acquire digitized law. 
In the decision-making process, the law-innocent RL agent will first generate smarter long-term RL action. 
The law-compliance of the RL action is checked with the digitized law by the law-violence forecaster. 
The sample-based backup policy will intervene to generate law-compliance action when the RL action is forbidden. 

The law-violence forecaster identifies law-violent RL actions before the law is broken, and evokes the backup policy to keep SV legal.
Based on traffic law digitization, it first converts laws to constraints on trajectories and is able to adapt to law change easily.
A long-term action space for the policies is then designed, which is necessary for advanced law-violence identification.
When the RL action is illegal, the forecaster will evoke the backup policy at an appropriate opportunity to keep the SV complying with traffic laws.

The long-term law-innocent RL agent is designed for decision-making under normal circumstances, where the variable laws are not broken.
The RL agent is trained with several common laws, e.g., traffic lights, but is innocent of the variable laws and the newly issued laws. 

The law-adaptive sample-based backup policy can adaptively generate law-compliant actions.
Several available target positions are first sampled, and a polynomial-based trajectory generator calculates candidate trajectories.
The unsafe or law-violent trajectories will be filtered then.
And the final trajectory is chosen by minimizing a cost function on the rest trajectories.

Through the framework, in common scenarios, the RL agent can make an optimized decision, which helps maintain the advantage of the well-trained RL agent. 
Thus, self-driving vehicles are able to efficiently pass through complex scenarios which the rule-based backup policy usually cannot handle.
When meeting variable laws to which the RL agent is hard to adapt, the backup policy will intervene in time to keep the vehicle compliant with the law.

\section{Traffic Law Digitization}
\label{seclawdig}
In this section, the MDP problem is first introduced.
The natural language described laws are then formulated as digital constraints on MDP using Linear Temporal Logic (LTL) method.

\subsection{Markov Decision Process}
The autonomous driving task is usually formulated as an MDP problem in RL literature.
An MDP problem is defined as a tuple $(\mathcal{S}, \mathcal{A}, r, \mathcal{P})$. where $\mathcal{S}$ denotes the state space, $\mathcal{A}$ denotes the action space, $r:\ \mathcal{S}\times \mathcal{A}\times \mathcal{S}' \rightarrow \mathbb{R}$ is the reward function, and $\mathcal{P}:\ \mathcal{S}\times \mathcal{A}\times \mathcal{S}' \rightarrow [0,1]$ denotes the transition probability function. Besides, a fixed value $\gamma$ is defined as the discount factor for future reward.\cite{cao2021confidence}.

A policy $\pi$ maps each state $s$ to a distribution on the action space $\mathcal{A}$, i.e., $\pi:\ \mathcal{S} \rightarrow \Pr(\mathcal{A})$.
The probability of applying action $a$ under state $s$ with policy $\pi$ is referred as $\pi(a|s)$.
This paper uses the deterministic policy $\pi$, i.e., $a = \arg \max_a \pi(a|s)$.
An MDP problem tries to solve the best policy to maximize the cumulative reward.

A trajectory $\tau$ is defined as a tuple of experienced states and applied actions, i.e.,
\begin{equation}
    \tau(s_0) = \left\{s_0, a_0, s_1, a_1, \cdots \right\}
\end{equation}
where $s_{t+1}$ is the next state after state $s_t$ and $a_t$, i.e., $s_{t+1} \sim \mathcal{P}(s|s_t, a_t)$. 
$\tau(s_0)$ indicates the trajectory starts from $s_0$.

A trajectory generated by a policy $\pi$ denotes the trajectory in which the actions are generated from $\pi$,
\begin{equation}
    \tau_\pi(s_0) = \left\{s_0, a_0^\pi, s_1, a_1^\pi, \cdots \right\}
\end{equation}
where $a^\pi_t \sim \pi(a|s_t)$.

In this work, the trajectory until timestamp $t$ is further referred as $\tau^t$, i.e.,  $\tau^t= \left\{s_0, a_0, s_1, a_1, \cdots s_t, a_t\right\}$. And the trajectory using RL policy $\pi_{rl}$ before timestamp $t$ and backup policy $\pi_b$ from $t$ is defined as $\tau_{\pi_{rl}}^t(s_0) + \tau_{\pi_b}(s_t)$, i.e., 
\begin{equation}
\begin{aligned}
\tau_{\pi_{rl}}^{t}\left(s_{0}\right)+\tau_{\pi_{b}}\left(s_{t}\right)=&\left\{s_{0}, a_{0}^{\pi_{rl}}, s_{1}, a_{1}^{\pi_{rl}}, \cdots, s_{t-1}, a_{t-1}^{\pi_{rl}}\right.\\
&\left.{s}_{t}, a_{t}^{\pi_{b}}, s_{t+1}, a_{t+1}^{\pi_{b}} \cdots\right\}
\end{aligned}
\end{equation}

\subsection{Traffic Law Digitization using Linear Temporal Logic}
\subsubsection{Linear Temporal Logic}
Linear Temporal Logic (LTL) is a widely used method for traffic law formulation \cite{rong2020safe, esterle2019specifications,rizaldi2017formalising}. 
LTL describes the logic property of a temporal trajectory.
LTL composes of several atomic propositions, Boolean operators (and$\vee$, or$\wedge$, not$\neg$, imply$\Rightarrow$) and basic temporal operators (next$X$, until$U$, always$G$, eventually$F$)\cite{esterle2019specifications}.

% The atomic propositions map from an instant state-action pair to a Boolean. 
An LTL formula $\varphi$ is defined as a composition of atomic propositions with their logical relationships, including Boolean and temporal connectives.
According to whether a trajectory meets the logic described in an LTL formula, the formula $\varphi$ maps a trajectory to a Boolean, which can be described as,
\begin{equation}
\label{ltl}
    \varphi:\ \mathcal{T} \mapsto \left\{0,1\right\}
\end{equation}
where $\mathcal{T}$ denotes the space of trajectory. $\varphi(\tau) = 1$ indicates a compliant trajectory $\tau$ (described as $\tau \models \varphi$ in LTL literature), whereas $\varphi(\tau) = 0$ indicates that the LTL formula is broken in trajectory.

\subsubsection{Traffic Law Digitization}
To identify law-violent actions, the road traffic laws described in natural language are first digitized using the LTL method.
The traffic law digitization process consists of logic analysis, LTL-based formulation and threshold chosen.

The logical analysis aims to decompose the traffic laws into atomic propositions and logical relationships between them. 
A number of traffic laws are described in or can be converted to a series of sub-laws, where each sub-law is described as \textit{when $p$, then $q$}.
Here $p$ denotes the trigger conditions, indicating when the traffic laws are triggered. And $q$ denotes the requirements to be met.
$p$ and $q$ can be further decomposed as formulas composed of atomic propositions and connectives.
Note that $p={\rm true}$ indicates that requirements $q$ have no trigger condition, i.e., $q$ always holds in the driving task.

Through logical analysis, the LTL formula can be established.
Road traffic laws should be always obeyed when the SV is driving.
And requirements $q$ should be met when the traffic laws are triggered, i.e., when $p$ holds. 
Thus, the LTL formula can be described as,
\begin{equation}
    \varphi = G(p\Rightarrow q)
\end{equation}
where the symbol $G$ indicates that the proposition \textit{always} holds in the trajectory.

Finally, there usually exists some qualitative threshold description in natural traffic laws, e.g., safe distance, move slowly, etc.
To choose  appropriate thresholds, the values can be assigned by experts or statistical data.

\subsubsection{Traffic Law Digitization Examples}

This work takes three examples to show the law digitization process. 
These three examples are chosen because they represent the typical types of traffic laws related to road infrastructures (Traffic Lights), temporary traffic control (Olympic Lanes), and driving behaviors (Overtaking).

\textbf{Example 1:} (Article 40 of \textit{Regulation on the Implementation of the Road Traffic Safety Law of the People's Republic of China}\cite{regulation}) The driveway signal lamps may give signals by: (2) red crossing light or red arrow, which means the vehicles along this lane are prohibited from proceeding. 

The LTL formula can be,
\begin{equation}
\begin{aligned}
    \varphi = G & \rm{(traffic\ light\ on\ ego\ lane\ \vee traffic\ light\ color\ red} \\ &\rm{ \Rightarrow \neg exceed\ stop\ lane)}
\end{aligned}
\end{equation}

The proposition on each state-action pair is that \textit{if there is a traffic light on the ego lane and the traffic color is red, the ego vehicle shall not exceed the stop lane.}

\textbf{Example 2:} (Article 37 of \textit{Road Traffic Safety Law of the People's Republic of China}\cite{law}) Where a special driveway is delimited on a road, only prescribed vehicles are allowed to pass within the special driveway, and no other vehicle may drive into the special driveway.

The LTL formula can be,
\begin{equation}
\label{lanelaw}
\begin{aligned}
    \varphi = G & \rm{(\neg (ego\ on\ special\ driveway \vee}\\& \rm{ego\  is\  not\ the\  prescribed\  kind))}
\end{aligned}
\end{equation}

\textbf{Example 3:} (Article 47 of \textit{Regulation on the Implementation of the Road Traffic Safety Law of the People's Republic of China}\cite{regulation}) When a motor vehicle overtakes another one ... and after there is a second necessary safe distance between them (ego and the lag vehicle), the overtaking vehicle shall turn on right turn light and return to the original lane.

The LTL formula can be,
\begin{equation}
\label{overtakelaw}
\begin{aligned}
    \varphi = G & (\rm{ego\ cross\ right\ driveway \Rightarrow} \\& \rm{ (duration\ of\ right\ light\ on\ge 3s}\\& \rm{ \vee (distance > safe\ distance)))}
\end{aligned}
\end{equation}

Different from Examples 1 and 2 which constrains each state-action pair, this law constrains a long-term SV trajectory. 
The historical information is required for checking the trigger (overtake) and the requirements (duration of light on).
Moreover, a number of traffic laws constrain the trajectory, e.g., laws about lane change, reverse, etc.

\section{Law-adaptive Hybrid RL Decision Making}
\label{secladm}
In this section, the proposed law-adaptive hybrid Reinforcement Learning decision-making method is introduced. 

\subsection{Law-violence Forecaster}
Based on the digitized law $\varphi$, the law-violence forecaster is designed to identify the law-violent actions in advance, and switch to the backup policy at an appropriate opportunity.
In this section, a long-term action space is first introduced for law-violence forecasting.
The opportunity for backup policy intervention is ensured available by carefully choosing the end of the long-term action.
Finally, the law-adaptive decision-making process is summarized and its effectiveness is clarified.

\subsubsection{Law-violence Forecast}
Identifying the law violence after the laws are broken is meaningless. 
Besides, a number of laws constrain a long-term trajectory of SV instead of an instant state-action pair, which brings difficulty to law-violence forecasting.
Thus, a long-term trajectory is necessary for law-violence identification, which requires a knowledge of future ego action and environment transition.
In this work, the ground-true future transition of the environment is supposed to be provided by an environment prediction module.
A long-term action space for the RL agent and the rule-based policy is designed for future ego action estimation.

To build a long-term action, the policy will generate a target position, then a trajectory generator will calculate a trajectory to this target. 
% Thus, the legality and comfort can be checked with the long-term trajectory.
The action space can be described as a set of available positions $p_d$ in a relative coordinate system.
The trajectory generated can be described as,
\begin{equation}
    \tau = g(s_0,p_d)
\end{equation}
where $s_0$ denotes the current state, and $g$ indicates the trajectory generator.
The detailed design of $p_d$ and $g$ is introduced later.

Thus, the legality of the SV can be checked for a future period by the digitized traffic law $\varphi$.

It should be noted that the generated long-term trajectory is only a possible trajectory in the future, and $\varphi(\tau) = 1$ indicates the SV can find a law-compliant trajectory until $p_d$, whereas $\varphi(\tau)=0$ only indicates that the RL agent may break the law in the future.
% Thus, the law-violence forecasting is a little conservative.

%that there exists an available trajectory for SV to comply with laws at least until $p_d$.
% But $\varphi(\tau) = 0$ does not denote that no law-compliant trajectory 

\subsubsection{Backup Policy Intervention}
To keep the SV compliant with the law, when the RL actions break the law, the backup policy should intervene at the appropriate opportunity when it is able to correct the SV.

An effective policy switch should be performed when the backup policy is able to comply with the law in the future, i.e.,
\begin{equation}
\label{sw}
    t_s \leq \max_t(\varphi(\tau_{\pi_{rl}}^t(s_0) + \tau_{\pi_b}(s_t))=1)
\end{equation}
where $t_s$ denotes the switch time.
For example, due to inertia, the policy switch when the SV is on the point of crossing the stop line cannot prevent law violence.
Thus, the backup policy must intervene before the SV can stop in front of the stop line.

% Besides, policy switch performed just on the capacity boundary of the backup policy is sometimes irrational. 
% For example, in the stop line case, the backup policy can prevent law-breaking when the distance to the stop line is longer than the braking distance with the maximum deceleration, but it causes harshness.
% Besides, the capacity boundary can be intractable with some complex scenarios and traffic rules.

To ensure the effectiveness of the backup policy, there should exist at least an available state for intervention in the long-term trajectory.
Thus, the SV can tracking the law-compliant RL trajectory until the available state.
In this work, the end of the trajectory is chosen.
Let $s_e(\tau)$ denotes the end state of long-term action $\tau$ (at position $p_d$) and $t_e$ the related timestamp, it should meet the following condition,
\begin{equation}
\label{pdcond}
    \varphi(\tau_{\pi_{rl}}^{t_s}(s_0)+\tau_{\pi_b}(s_{t_e})) = 1, {\rm if\ } \varphi(\tau_{\pi_{rl}}^{t_s}(s_0)) = 1
\end{equation}

To achieve \eqref{pdcond}, the destination position (i.e., the action space of policies) must be carefully designed. 
However, the boundary of the set of available positions is also usually intractable. 
But for most laws, when the SV is in the lane, and the lateral speed is 0 with legal longitudinal speed, the backup policy can keep it complying with the laws. 
And through our case study, this action space performs well. 

\subsubsection{Law-adaptive Decision-Making Process}
To summarize, the long-term action space is designed for both policies in advance. 
The action space $\mathcal{A}$ is a (sub)set of available target positions.

At each timestamp $t$, the RL agent first generate an optimized action $a_{rl} = p_{d1}$.
Then the forecaster will check the legality of the action by its trajectory $\tau_{rl}=g(s_0,p_{d1})$.
When it breaks the law, i.e., $\varphi(\tau_{rl})=0$, the backup policy will generate its action $a_b = p_{d2}$ and related trajectory $\tau_b = g(s_0,p_{d2})$.

However, at states during the RL trajectory, the effectiveness of rule-based policy is not ensured.
Thus, an additional trajectory buffer $p_{bf}$ is designed to store the last available trajectory.
It will be tracked until at least one policy is available.

Finally, the law-adaptive decision-making process can be described as,
\begin{equation}
    a^*_t = 
    \begin{cases}
    a_{rl} &  \varphi(\tau_{rl})=1 \\
    a_b & \varphi(\tau_{rl})=0 \ \rm{and} \ \varphi(\tau_{b})=1 \\
    a_{bf} & \varphi(\tau_{rl})=0 \ \rm{and} \ \varphi(\tau_{b})=0
    \end{cases}
\end{equation}
where $a_{bf}$ is the buffered target position. The buffer will be updated when RL action or backup action is available, i.e., 
\begin{equation}
    a_{bf} = a^*_t, \rm{when}\  \varphi(\tau_{rl})=1 \ \rm{or} \ \varphi(\tau_{b})=1
\end{equation}

The effectiveness of this method can be easily proved as follows.
When the RL action obeys the law, it will be performed.
When it breaks the law, the SV will try to evoke the backup policy.
If $a_b$ complies with the law, it will be performed. When it is illegal, the buffer action will be performed, and the SV legality is ensured by the buffer action.
Besides, since the target positions of RL and backup policy meet \eqref{pdcond}, the backup policy always can find an available timestamp to maneuver the policy switch before the buffer runs out.

The smoothness of the SV is pursued by the polynomial-based trajectory generator, and a controller is designed for tracking the final trajectory. 

\subsection{RL Agent Generation}
In this work, the deep Q learning\cite{mnih2013playing} framework is utilized for Reinforcement Learning agent training, while other RL frameworks may also work but are not validated.
Deep Q learning is an RL framework with discrete action space. It uses a deep neural network to estimate the value of action $a$ under state $s$. And the final action $a^*_{rl}$ is chosen by maximizing the expected value.

The long-term action space is bulit as,
\begin{equation}
    a = p_d = [p_{lat}, p_{lon}]\in \mathcal{A}
\end{equation}
where $p_{lat}\in\left\{-1,0,1\right\}$ denotes the target lane index. $-1, 0,1 $ indicate the left, ego, right lane, respectively. $p_{lon}\in[-1,1]$ denotes the normalized longitudinal acceleration.
The action space is further discretized for deep Q learning by discretizing the longitudinal acceleration.

In this work, the method is validated in cases with only one surrounding vehicle. 
The state is designed as,
\begin{equation}
    s = [q_e, q_o]
\end{equation}
where $q=[x,y,\dot x, \dot y]$ indicates the position and speed of the vehicles. 
Subscript $e$ and $o$ denote the ego and  the other vehicle, respectively.

The reward design depends on specific cases.
The RL agent is trained without the variable laws, but some common laws are designed in training, e.g., traffic light.
\subsection{Rule-based Backup Policy}
This work adopts a law-adaptive sample-based policy as the backup policy. Other traffic law adaptive policies can also be used but not verified in this work.

The policy first samples several available target positions from $\mathcal{A}$ as $p_{d1}, p_{d2}, \cdots, p_{dn}$, where $n$ denotes the number of samples.
For each $p_d$, a candidate trajectory is generated by the trajectory generator.

For each target position $p_{d}$, the trajectory generator will first convert it to world coordinate system. Then the trajectories are generated by a cubic polynomial-based trajectory generator as,
\begin{equation}
 \begin{aligned}
\vec{p}(t)=&\left(2 t^{3}-3 t^{2}+1\right) \vec{p}_{0}+\left(t^{3}-2 t^{2}+t\right) \vec{m}_{0} \\
&+\left(-2 t^{3}+3 t^{2}\right) \vec{p}_{1}+\left(t^{3}-t^{2}\right) \vec{m}_{1}, \quad t \in[0,1]
\end{aligned}   
\end{equation}
where $\vec{p}(t)$ denotes the points on the curve. $\vec{p}_{0}=$ $\left[x_{0}, y_{0}\right]^{T}, \vec{p}_{1}=\left[x_{1}, y_{1}\right]^{T} \quad$ denote the current position and desired position, respectively. $\vec{m}_{0}=\left[\dot{x}_{0}, \dot{y}_{0}\right]^{T}, \vec{m}_{1}={\left[\dot{x}_{1}, \dot{y}_{1}\right]^{T}}$ denote the ego vehicle starting velocity and desired ending velocity, respectively.\cite{cao2021confidence}
The long-term action is a combination of $\vec{p}(t)$ and environment transition acquired from the prediction module.

For the $n$ candidate trajectories, the unsafe trajectories will be filtered first through the prediction of other road users\cite{cao2021lidar}. 
Then the illegal trajectories will be identified with the aforementioned law-compliance forecaster. 
A cost function\cite{cao2021lidar} considering the smoothness is applied then for choosing the best trajectory.
The final trajectory can be referred as,
\begin{equation}
    \tau_b = \arg \min_\tau C(\tau) | \varphi(\tau) = 1, \ \varphi_s(\tau) = 1
\end{equation}
where $C$ denotes the cost function, and $\varphi_s(\tau) = 1$ indicates that trajectory $\tau$ is safe.

\section{Case Study}
\label{case}
In this section, a law-updating case and a temporary law case are used to verify the proposed method.
Both are typical urban driving scenarios built in the CARLA simulator\cite{Dosovitskiy17}.

\subsection{Beijing Winter Olympic Lane}
\subsubsection{Evaluation Scenario Setting}

\begin{figure}[htbp]
    \centering
    \subfigure[Winter Olympic Lane]{
        \includegraphics[trim=0 30 0 40, clip, width = 0.97\linewidth]{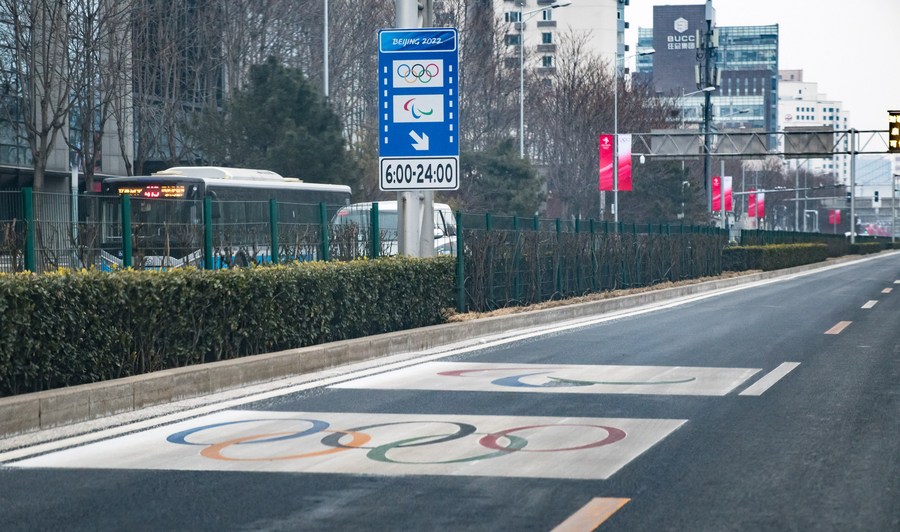}
    }
    \subfigure[Evaluation Scenario Setting of Winter Olympic Lane]{
        \includegraphics[width = 0.97\linewidth]{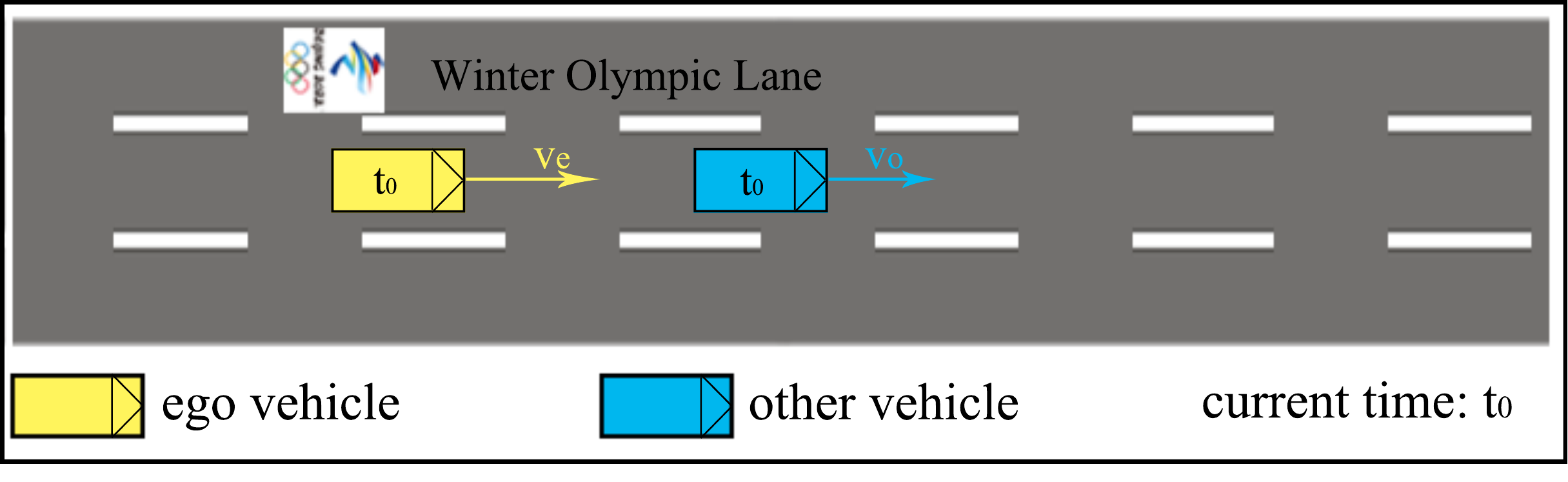}
    } 
    \caption{The Beijing Winter Olympic Lane Scenario.}
    \label{case2}
    % \vspace{-0.02\linewidth}
\end{figure}

The 2022 Winter Olympic Games are held in Beijing.
During the Event, several lanes are set up for exclusive use by Beijing 2022 Olympic and Paralympic Winter Games participants, and private vehicles are banned from using the lane as normal.
The traffic laws about special lanes are shown in \eqref{lanelaw}.

The scenario is shown in Fig.\ref{case2}. 
In this scenario, the ego vehicle drives on a three-way road, and the leftmost lane is set as the Event Lane.
The vehicle aims to pass through the area near a recommended speed.
Another vehicle drives slowly in front of the ego vehicle.
Thus, the ego vehicle may perform a lane change maneuver to keep its speed.

As a temporal traffic rule, the special lane is not considered in RL agent training. 

\subsubsection{Experiment Results}
\begin{figure}[htbp]
    \centering
    \subfigure[Law-violent RL action without the proposed method.]{
        \label{winterrl}
        \includegraphics[width = 0.98\linewidth]{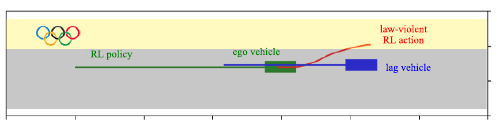}
    }
    \subfigure[The decision-making process of the sample-based backup policy.]{
       \label{wintersa}
       \includegraphics[width = 0.98\linewidth]{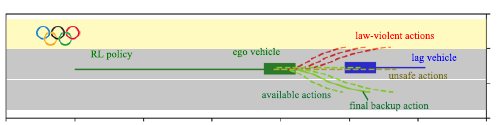}
    } 
    \caption{The ability of the proposed method to adapt to newly issued laws.}
    \label{res1}
\end{figure}

The experiment result is shown in Fig.\ref{res1}.
The SV action before and after adopting the proposed method are shown in Fig.\ref{winterrl} and Fig.\ref{wintersa}, respectively.
In this case, the RL agent tends to perform a left lane-change maneuver when approaching the lag vehicle to keep its speed.
Thus, without the proposed method, the SV will drive into the Event Lane illegally due to the innocent RL agent of the temporary traffic law as in Fig.\ref{winterrl}. 

By adopting the law-adaptive decision-making framework, the law-violent RL action is prevented by the forecaster, and the backup policy intervenes in time.
The detailed decision-making process of the backup policy is shown in Fig.\ref{wintersa}. 
Totally eight trajectories of sampled targets are generated.
The illegal trajectories (in red dashed lines) and unsafe trajectories (in yellow dashed lines) are first filtered. 
Then the trajectory with minimum cost (in this work, near the recommended longitudinal speed) is chosen as the final backup policy.
Thus, the SV turns to the right lane and passes through the Event Lane area legally.

The results indicate that the proposed method can help SV adapt to new issued or temporary road traffic laws, which can also deal with emergency traffic control or road closure.

% LTL check RL 
% sample-based rule chosen

\subsection{Safe Distance in Overtaking}
\subsubsection{Evaluation Scenario Setting}
\begin{figure}[htbp]
    \centering
    \includegraphics[width = \linewidth]{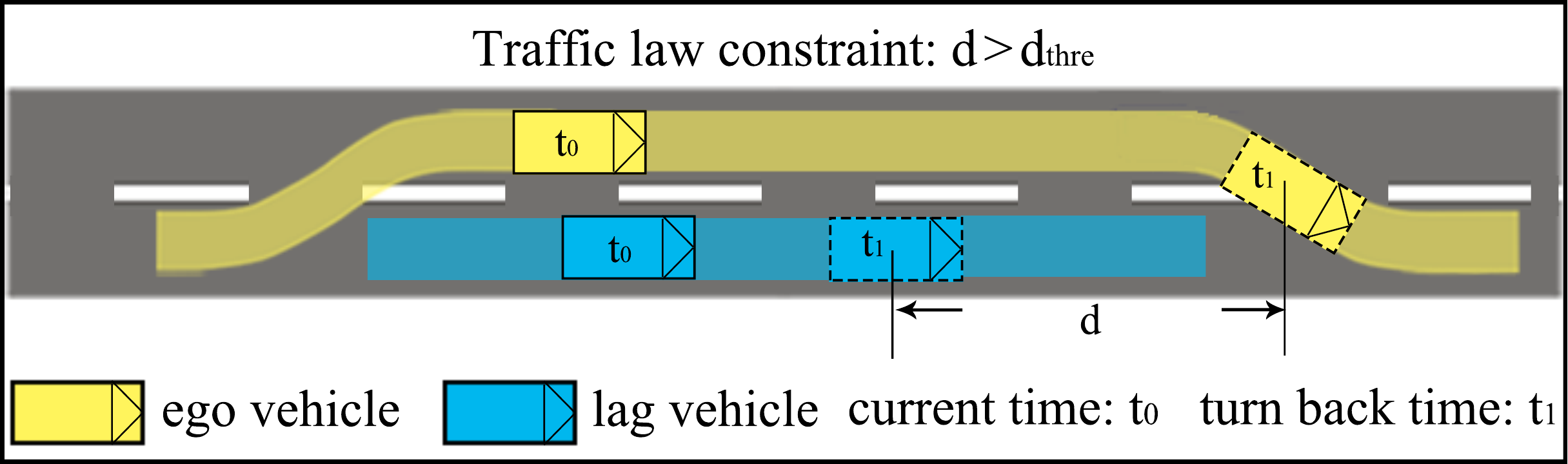}
    \vspace{-0.2in}
    \caption{Evaluation Scenario Setting of Overtaking Scenario}
    \label{casedesign}
\end{figure}

In this case, the method is tested in a typical overtaking scenario on a two-way road, as shown in Fig.\ref{casedesign}.

It describes a part of overtaking maneuver.
In this scenario, the ego vehicle tries to overtake the lag vehicle from the left lane. 
The left lane changing and accelerating maneuver has finished, and the ego vehicle tries to change back to the origin lane now.
% The lag vehicle will always keep in lane, and its longitudinal speed keeps constant.

The related traffic law is shown in \eqref{overtakelaw}.
In this case, we only focus on the turn-back distance.
And the turn-back time is defined as when the ego vehicle crosses the lane line.

\begin{table}[htbp]
    \centering
    \caption{Different Traffic Law Setting}
    \renewcommand{\arraystretch}{1.3}
    \begin{tabular}{c|c}
    law type & distance threshold value  \\
    \hline
    fixed distance & $d_{thre} = d_{min}$ \\
    variable distance & $d_{thre} = d_0 - \tau\Delta v $\\
    mixed distance & $d_{thre} = \max \left\{d_{min}, d_0 - \tau\Delta v  \right\}$ 
    \end{tabular}
    \label{lawsetting}
\end{table}
 
To test the ability of our method to adapt to different laws, three kinds of the safe distance threshold are set, shown in Table.\ref{lawsetting}, where $d_{thre}$ denotes the threshold. $d_{min}$ is the fixed distance threshold, set as $12$m. $d_0$ and $\tau$ are the reference distance and time headway reference for variable distance threshold, respectively, set as $d_0 = 18$m and $\tau = 1.5$s. $\Delta v$ denotes the relative speed of ego vehicle to the lag, i.e., $\Delta v = v_e - v_o$, where $v_e$ and $v_o$ denote the longitudinal speed of ego and lag vehicles, respectively.
The mixed distance threshold is a combination of the forward two criteria.
% The distance threshold is low since the vehicles drives slowly in this case, which is limited by the road length in CARLA simulator. But the following results are also available for high speed scenarios.

\begin{figure}[htbp]
    \centering
    \subfigure[Fixed distance threshold]{
        \label{res11}
        \includegraphics[width = 0.97\linewidth]{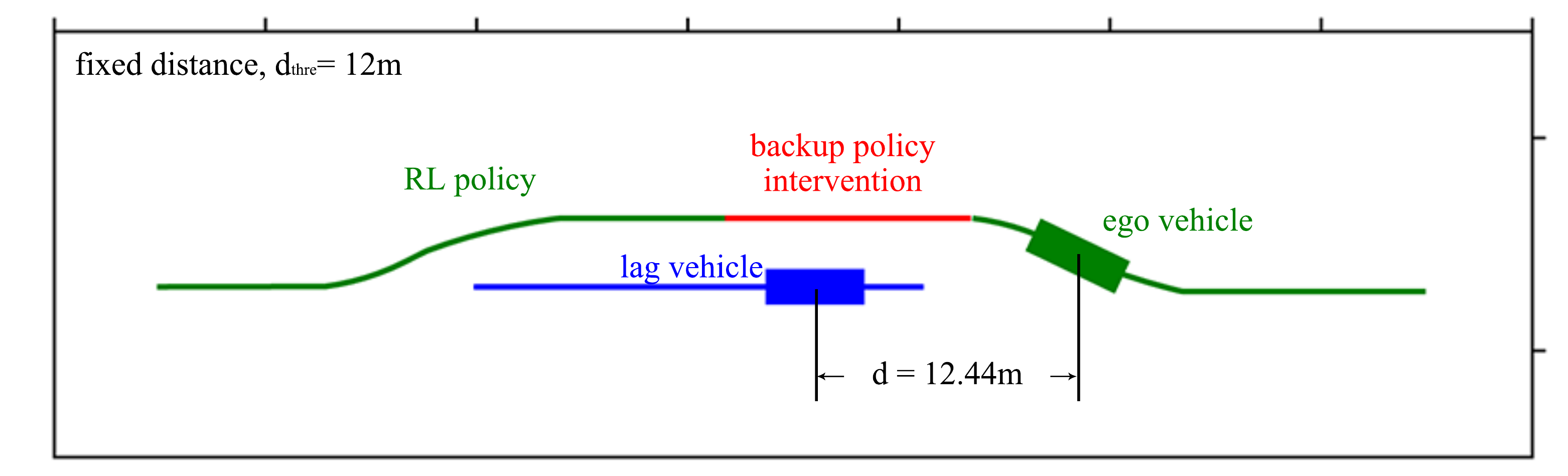}
    }
    \subfigure[Variable distance threshold]{
       \label{res12}
       \includegraphics[width = 0.97\linewidth]{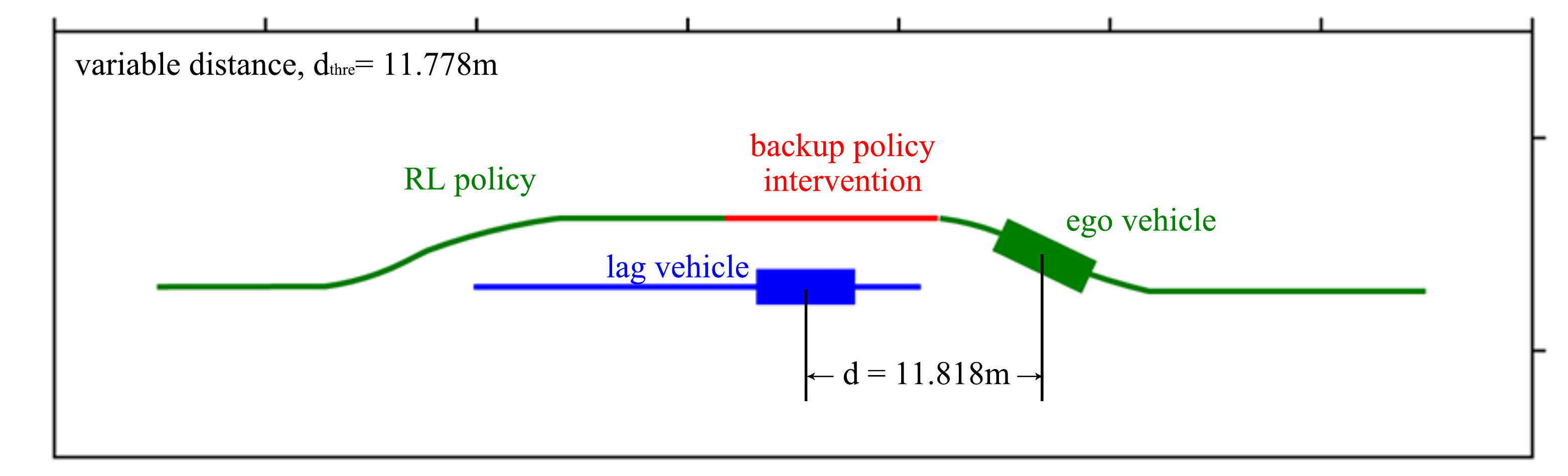}
    } 
    \subfigure[Mixed distance threshold]{
        \label{res13}
        \includegraphics[width = 0.97\linewidth]{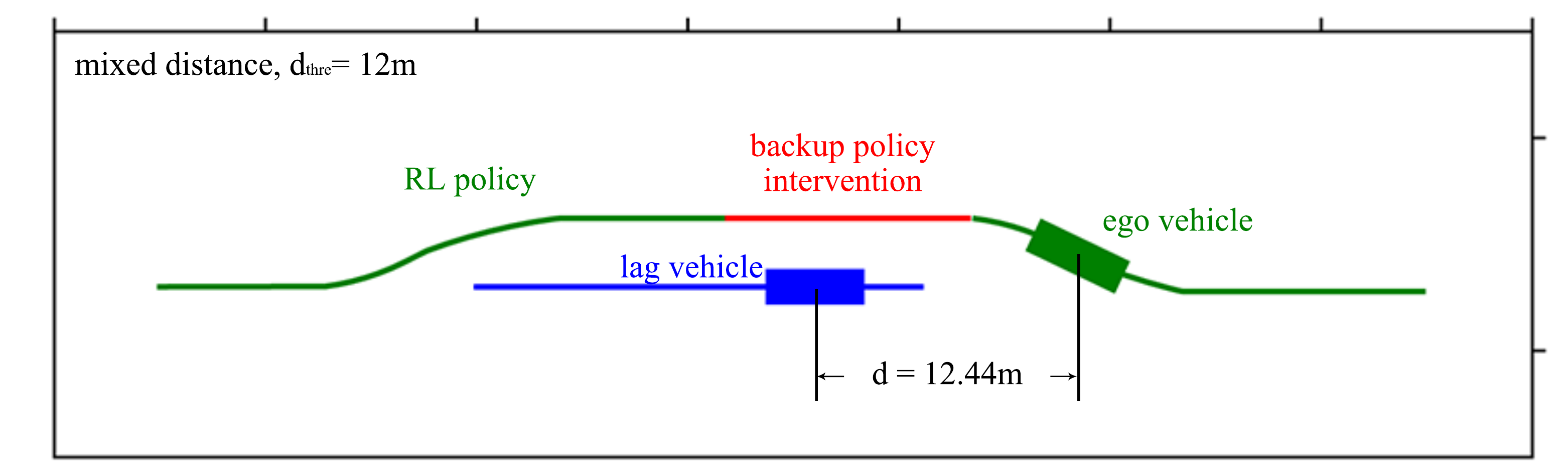}
    }
    \caption{The ability of the proposed method to adapt to different laws.}
    \label{result1}
\end{figure}

\subsubsection{Experiment Results}
The experiment result is shown in Fig.\ref{result1}.
The SV performance under fixed, variable and mixed distance threshold of the overtake traffic law is shown in Fig.\ref{res11}, \ref{res12} and \ref{res13}, respectively.
In this scenario, the RL agent makes the turn-back decision when the maneuver is safe but breaks the law.
The law-violent actions are identified by the forecaster, and the backup policy intervenes to keep the SV driving straightforward (the red line in Fig.\ref{result1}), which complies with the traffic laws.

The results indicate that the proposed method can adapt to law updating well, which is also necessary for SV to deal with different explanations and thresholds chosen for the same road traffic laws.

\section{Conclusion}
\label{seccon}
In this work, a hybrid reinforcement learning decision-making method was proposed to make self-driving vehicles adapt to variable road traffic laws without the demand for abundant driving data and laborious re-training.
The main idea is to supply the defect of the RL agent which lacks the adaptability to traffic laws modification by a sample-based law-adaptive backup policy.
The laws described in natural language are first converted to constraints on Markov Decision Process by the Linear Temporal Logic method.
A law-violence forecaster identifies law-breaking actions in advance based on a long-term action space.
A deep Q learning-based RL agent is designed as the backbone and a sample-based backup policy will intervene to re-plan a law-compliant trajectory when the RL action breaks the traffic laws.
Through the framework, the law modification adaption of self-driving vehicles can be easily conducted in the forecaster and the backup policy without re-training and re-collecting data.
Through validation in a Beijing Winter Olympic Lane case and an overtaking case built in the CARLA simulator, the results showed that our method is able to adapt to updated and new-issued laws effectively.
For self-driving vehicles, this method provides a manner for vehicle managers to govern them, e.g., manufacturers or governments, which is necessary for the wide landing of autonomous driving.

\section*{Acknowledgment}
	The authors would like to appreciate the financial support of the National Science Foundation of China Project (52072215, U1964203, 52102460) and National key R\&D Program of China (2020YFB1600303).
	
\bibliographystyle{IEEEtran} %声明选择的格式
\bibliography{ref}

\end{document}